\documentclass{article}
\usepackage[english, norsk]{babel}
\usepackage[utf8]{inputenc}
\usepackage{todonotes}
\usepackage{mathtools}
\usepackage{amsmath}

\usepackage{authblk}			% author block front matter
\usepackage{graphicx}			% images
\usepackage{subcaption}
\usepackage{siunitx} 			% units
\usepackage{amsmath} 			% matrices
\usepackage{amsfonts} 			% mathbb
\usepackage{biblatex}
\bibliography{main.bib}
\usepackage{indentfirst}
\usepackage[capposition=bottom]{floatrow}
\usepackage[hidelinks]{hyperref}
\hypersetup{
    colorlinks=true,
    linkcolor=blue,
    filecolor=magenta,      
    urlcolor=black,
    citecolor=blue,
    pdftitle={Overleaf Example},
    pdfpagemode=FullScreen,
    }
    
\makeatletter
\renewcommand{\maketitle}{\bgroup\setlength{\parindent}{0pt}
\begin{flushleft}
  {\LARGE \textbf{\@title}\vspace{1em}}

  {\@author}
 
\end{flushleft}
\begin{flushright}
    Trondheim, Norway, \today
\end{flushright}
\egroup
}
\makeatother
\renewenvironment{abstract}
 {\par\noindent\textbf{\abstractname}\ \ignorespaces}
 {\par\medskip}
\usepackage[margin=1.0in]{geometry}

\title{Underwater autonomous mapping and characterization of marine debris in urban water bodies}
\author[1]{Trygve Olav Fossum}
\author[1]{Øystein Sture}
\author[1]{Petter Norgren-Aamot}
\author[2]{Ingrid Myrnes Hansen}
\author[3]{Bjørn Christian Kvisvik}
\author[4]{Anne Christine Knag}
\affil[1]{\textit{Skarv Technologies AS, contact@skarvtech.com}}
\affil[2]{\textit{Ecotone AS, info@ecotone.com}}
\affil[3]{\textit{COWI AS}}
\affil[4]{\textit{City of Bergen, Department of Climate, Environment and Urban Development, Agency for Urban Environment}}

\begin{document}

\maketitle

%\section*{}
\selectlanguage{English}

\begin{abstract} 
Marine debris originating from human activity has been accumulating in underwater environments such as oceans, lakes, and rivers for decades. The extent, type, and amount of waste is hard to assess as the exact mechanisms for spread are not understood, yielding unknown consequences for the marine environment and human health. Methods for detecting and mapping marine debris is therefore vital in order to gain insight into pollution dynamics, which in turn can be used to effectively plan and execute physical removal. Using an autonomous underwater vehicle (AUV), equipped with an underwater hyperspectral imager (UHI) and stereo-camera, marine debris was autonomously detected, mapped and quantified in the sheltered bay Store Lungegårdsvann in Bergen, Norway. 
\end{abstract}

\section{Introduction}
Marine pollution is a problem that has been building for decades and poses a substantial threat to marine wildlife and ocean health \cite{national2021reckoning}. The pollution encompass everything from plastic waste and abandoned fishing gear, to industrial and agricultural discharge. Impacts to marine organisms can be difficult to quantify directly, but are well known. Ingested marine debris, particularly plastics, has been reported in necropsies of birds, turtles, marine mammals, fish, and squid \cite{board2009tackling}. This anthropogenic impact is especially high in city areas close to the ocean with high population density. Particularly open ocean plastics have been shown to increase in recent decades \cite{ostle2019rise}. The debris is
transported into larger water bodies by rivers, winds, sewers, and accumulates based on local tide and current conditions. Mapping and monitoring of marine pollution is therefore critical for understanding the extent and patterns of waste input to the environment. The physical removal of coastal marine debris is a costly and time consuming activity, which needs to be performed using specialized tools and vessels. \vspace{1em}

By leveraging smaller autonomous underwater vehicles (AUV) and different optical instruments, such as UHI and stereo vision, we demonstrate a methodology for doing effective mapping and detection of marine debris on the seabed, as well as characterization of sediment types. The survey is conducted in the sheltered bay Store Lungegårdsvann in Bergen, Norway. The AUV platform used is developed by Skarv Technologies\footnote{\url{www.skarvtech.com}} specifically for light weight seabed mapping and can operate both as an AUV and ROV, hence the abbreviation (AROV). The underwater hyperspectral imager (UHI) sensor is developed by Ecotone\footnote{\url{www.ecotone.com}} AS for \emph{close-range remote sensing} applications, such as subsea inspection and environmental mapping in turbid waters. The rest of this paper is organized as follows. We describe previous work in Section 2. In Section 3 we give a overview of the project and survey area, the survey platform, and the methods involved in automatic detection. Section 4 contain data-analysis and results from the field survey, followed by a discussion and conclusion in Section 5 and 6.

\section{Related Work} \label{sec:related}
Marine debris detection using acoustic data has been explored in \cite{valdenegro2016submerged}, based on laboratory recordings with a forward looking sonar and a deep convolutional neural network (CNN). Using video images collected from the Japan Agency for Marine Earth Science and Technology JAMSTEC E-library of Deep-sea Images (J-EDI) data set (J. A. 2012) , \cite{fulton2019robotic} demonstrates automatic detection of marine debris using several deep visual detection models. The work demonstrates that detection of marine litter is possible (even for real time applications), but limited by the large pattern variability intrinsic to the objects and underwater environment. Accompanying the paper is also the "TrashCan 1.0" labelled data set \cite{TrashCan}, which is publicly available. Building on this data set and employing the EfficientDet \cite{Wightman} model, \cite{zocco2022towards} explores the use of CNNs in real-time and low-light conditions, successfully employing methods to enhance images before detections. The paper also includes the in-water plastic bags and bottles (WPBB) data set. A more broad approach to trash detection is taken in \cite{MAJCHROWSKA2022274}, which included training on data sets from both terrestrial (indoor and outdoor) and underwater environments, attaining accuracies up to 75\%. The merged collection contains over 28'000 images and over 40'000 objects, providing much broader data diversity than previous approaches. Diversity is important to account for the wide range of debris, degradation, morphology, and environmental conditions that exists especially for marine debris. Attaining good predictions for debris that has been in the ocean for a long time, is broken down to smaller pieces, and is partly covered by sediments continues to be a challenge. Hence, the use of optical instruments, like the UHI, can be used to augment these detection systems by helping to discriminate smaller objects and providing additional information about the material properties. 

\vspace{1em}

Airborne hyperspectral remote sensing for the purpose of environmental monitoring of land, coastal areas and shallow water habitats, is an  established technology \cite{stuart2019hyperspectral}. During the last decade, hyperspectral imaging for \emph{underwater} use has emerged as an important tool for both science and industry with the  instrumentation integrated on a wide range of underwater vehicles, such as remotely operated vehicles (ROV) \cite{johnsen2016use, rs14061325}, unmanned surface vehicles (USV) \cite{mogstad2019shallow} and autonomous underwater vehicles (AUV) \cite{sture2017autonomous}. Much of the research on hyperspectral imaging underwater, has focused on new and improved methods for subsea environmental monitoring applications. Today, UHI for  for material identification and environmental mapping is well documented \cite{liu2020underwater,montes2021underwater}. UHI operated from underwater vehicles has been used to characterize cold-water coral habitats \cite{mogstad2022remote},to assess deep sea megafauna \cite{dumke2018underwater}, to map the environmental footprints from offshore drilling operations \cite{cochrane2019detection} and to monitor the extent of organic wastes from aquaculture production \cite{Salmar}. The UHI instrument needs to compensate for water attenuation (inherent optical properties of the water), and attenuation of light based on distance to target, in order to give comparable measurements of the \emph{spectral reflectance} or just \emph{reflectance}. Reflectance is the unique response between light and material, which is used to identify the material type. Generally plastics are best separated spectrally in the near-infrared (NIR) and mid-infrared spectrum (900-1800nm), see e.g. \cite{Kumagai02}. Detection of plastics objects is however a simpler task that possible to approach using the visible spectrum (400-700nm). The UHI instrument used in this survey is developed by Ecotone AS$^2$ and is a push-broom type imager that covers the spectrum from 380nm to 750nm, which makes it a relevant tool in the context of marine debris detection.

\section{Materials and Methods}
\subsection{Store Lungegårdsvann and the "Renere Havn Bergen" project}
The study area is the sheltered bay Store Lungegårdsvann (60$^{\circ}$22'54.7"N 5$^{\circ}$20'38.1"E) outside the city of Bergen. The seabed in the shallow parts is characterised by fine-coarse sand, river sediments, and rocky shoals, while deeper parts of the bay is oxygen depleted with bacterial/microbial mats covering the seabed. Large parts of the seabed outside Bergen are heavily polluted by marine debris, heavy metals, and organic pollutants. The aim of "Renere Havn Bergen" project \footnote{\url{https://www.bergen.kommune.no/hvaskjer/tema/renere-havn-bergen}} is to prevent the spread of environmental toxins from the polluted seabed and to create a cleaner port environment. The project is responsible for investigations of the environmental condition, testing of new technologies, and implementation of clean-up processes. The intervention area is a total of approx. 3 km$^2$ and covers Puddefjorden, Store Lungegårdsvann, and Vågen. The measures against polluted seabed in the port of Bergen are among the largest, most expensive and most complicated environmental projects in the west of Norway.

\subsection{AROV-SKR300 - An autonomous ROV system for advanced optical surveys}
\begin{figure}[!h] 
\centering
\includegraphics[width=0.85\textwidth]{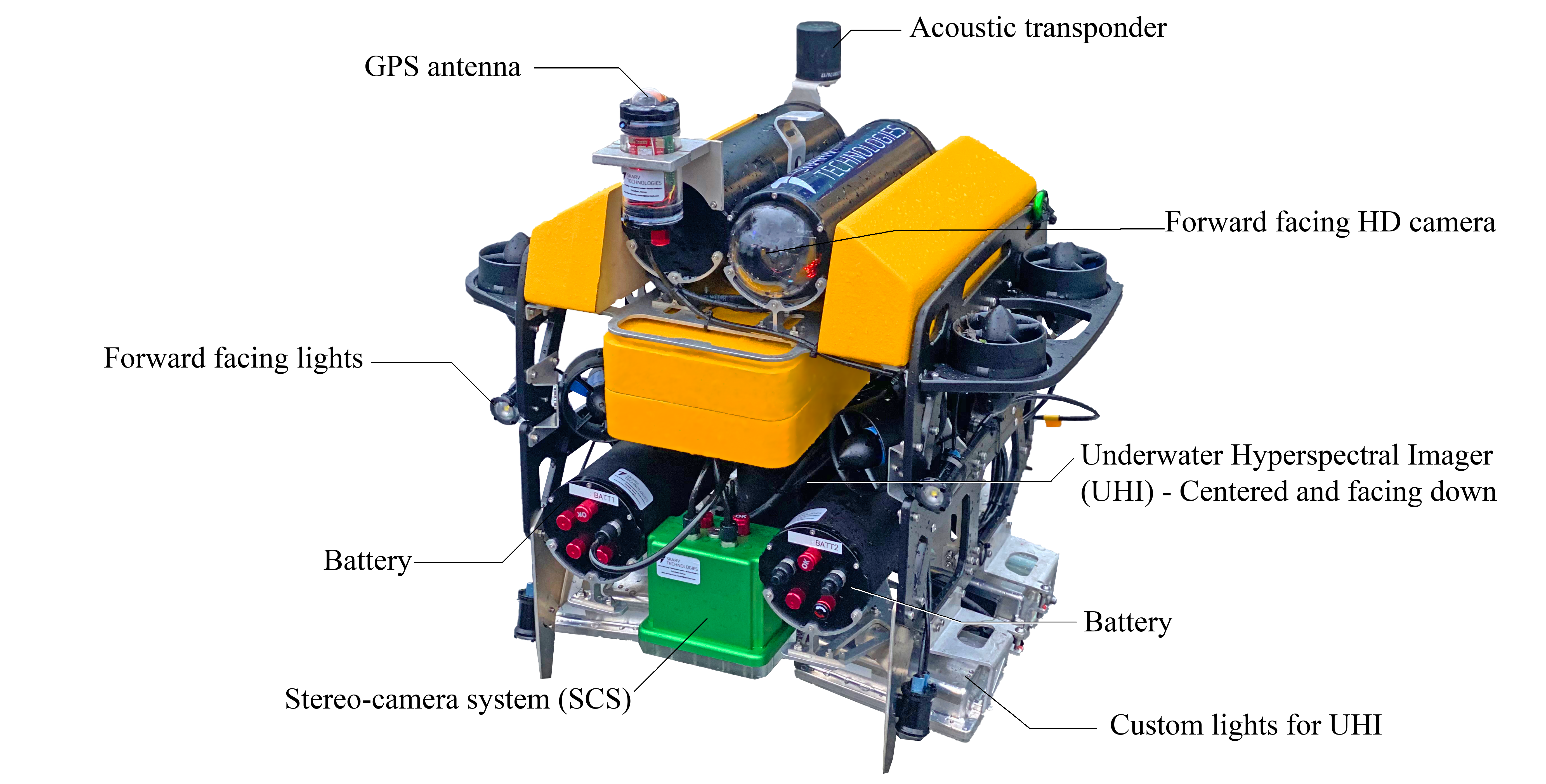}
\caption{The AROV (SKR-300) platform.}
\label{fig:AROV}
\end{figure}

The autonomous-ROV (AROV) system (see Figure \ref{fig:AROV}) is the among the first compact systems specially designed for high resolution optical seabed inspection and mapping, capable of both remote and fully autonomous operation (tether can be dis-attached). A proprietary underwater navigation system combining an inertial measurement unit (IMU), Doppler velocity log, and stereo-camera system (SCS) enables operations to be conducted autonomously without any topside link. Acoustic communications can be added to maintain low bandwidth communications and provide position updates (ultra short baseline - USBL). The main payload sensors are the SCS and the UHI instrument, supported by custom subsea lights that are optimized for turbid water conditions. The AROV system is based on a highly modified BlueROV2 (heavy configuration) \cite{robotoics2021bluerov2} from the company BlueRobotics\footnote{\url{https://bluerobotics.com/}} and can be configured to carry multiple sampling, manipulator, and observational tools depending on the application. The modular base-design leveraging off-the-shelf parts allows the user to add and remove components depending on the operational requirements without significant cost. Onboard processing of sensor data, such as video and images, is supported and can be used to trigger reactive and adaptive mission capabilities (obstacle avoidance, hovering, target tracking, etc.). The AROV system is compact (35kg) and can be operated by a single person through a mission and commanding interface. The central specifications of the AROV system is listed in Table \ref{tab:AROV_parameters}.

\begin{table}[!h]
\caption{Technical specifications of the AROV (SKR-300) system. Optional$^*$}
\begin{tabular}{lll}
\textbf{AROV-SKR300 Parameters} & \textbf{Value}\\
Physical dimensions and weight & Height x Width x Length $58cm \times 52cm \times 58cm$, $45$kg \\
Maximum depth & $100-300$m (depending on configuration)\\ 
Speed & 2 knots (max speed) \\ 
Battery endurance & 10-20 hrs (depending on configuration) \\ 
Optical sensors & Front facing camera, stereo-camera, UHI*\\
Other sensors & GPS, DVL, Subsea lights (custom) \\ 
Communications & WLAN, Acoustic Modem* (EvoLogics)\\ 
\end{tabular}
\label{tab:AROV_parameters}
\end{table}

\subsection{Automatic object detection and classification}
\begin{figure}[!h] 
\centering
\includegraphics[width=0.99\textwidth]{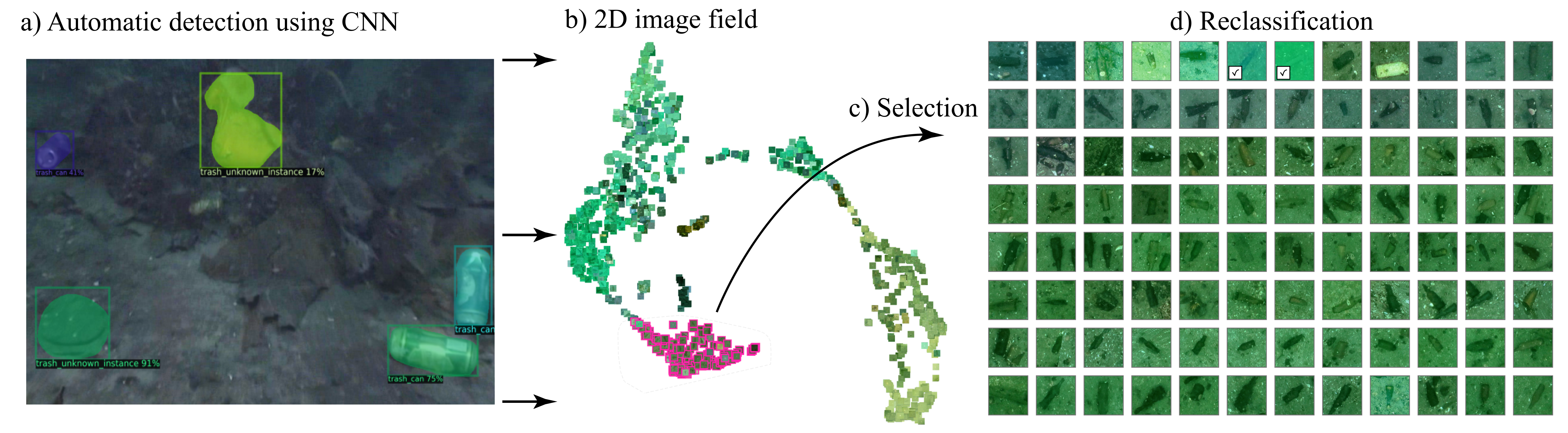}
\caption{Example figure explaining the steps involved using the CVI-tool for checking and verifying the automatic detections (here seen from the instance class "bottles") from the CNN. a) CNN example detecting different marine debris, b) 2D image field clustered according to pattern and spectral features, c) Selection of images from image field, and d) reclassification for erroneous detections within the selection.}
\label{fig:image_sorting}
\end{figure}

The proposed detection and classification framework is split into  i) an automatic detection system (ADS), and  ii) a semi-automatic clustering and visualization (CVI) tool to validate the classification results. The two parts combine RGB-image detections (pattern-based), human verification, and UHI-images (spectral information) to attain a high level of accuracy and confidence required for subsequent analysis and quantification. The automatic object detection system (i) is based on the detection and segmentation network Mask R-CNN \cite{mask-r-cnn}. The data set used for training the ADS consisted of $>$1000 manually annotated RGB-images from the survey area captured on the SCS, subject to data augmentation to increase the size of the training set further. The ADS  represents a proof-of-concept for future work in detecting marine debris and is not optimized for accuracy or performance. Given the probability of high turbidity, partly buried marine debris, and limited training data (representing turbid conditions), the accuracy of the automatic detection was expected to be low (40-60\%). Thus a secondary refinement procedure was added to achieve acceptable results that could be used for planning removal operations of the marine debris. A semi-automatic clustering and visualization (CVI) tool (ii) was therefore devised to easily check, reclassify, and possibly remove erroneous detections. The CVI-tool allows inspection and validation of individual detections as image elements, as well as co-registration of UHI-data; where co-registration entails georeferencing together with the stereo-images. Figure \ref{fig:image_sorting} illustrates a typical workflow, were the automatic system (i) feeds detections into the CVI-tool (ii). Similar detections are shown to a human inspector in a two dimensional (2D) \emph{image field} (see Figure \ref{fig:image_sorting}b) based on feature similarity (pattern and spectrum [gathered from the UHI-recording]) and class uncertainty. The inspector would then select (Figure \ref{fig:image_sorting}c), sort out, or potentially reclassify (Figure \ref{fig:image_sorting}d) any mis-classified detections by exploiting the image field. Several different views can be toggled to sort the images according to different feature vectors, such as: pattern, spectrum, and classification probability. More intricate details of the system is considered intellectual property of Skarv Technologies and is thus not exhaustively covered here.    

\section{Results from field survey in Store Lungegårdsvann}
\begin{figure}[!h] 
\centering
\includegraphics[width=0.99\textwidth]{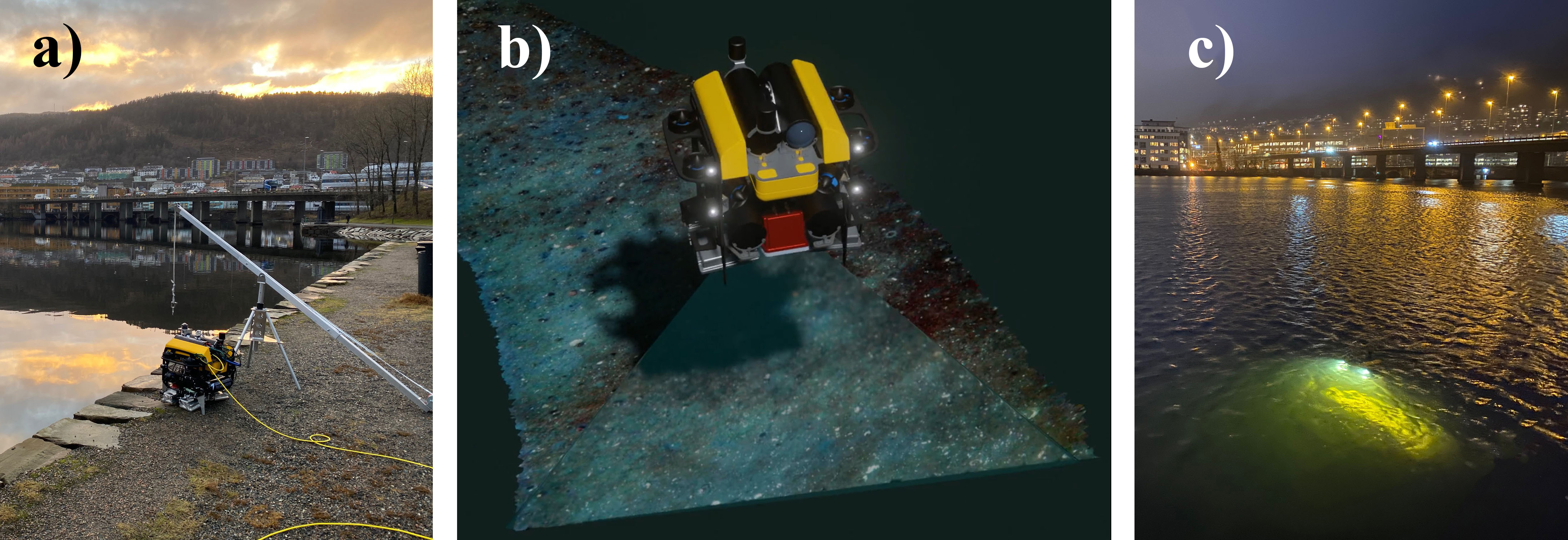}
\caption{a) Mobile crane and deployment setup. b) Survey illustration of the data collection methodology using the AROV system. c) Survey in progress during night. The specialized UHI lights (for turbid conditions) can be seen as a focused line.}
\label{fig:survey}
\end{figure}

Two one-week survey periods were conducted with the AROV-system; one taking place in September 2021 and the other in January 2022. A total of 70'000 $m^2$ was mapped and subseqently analysed to determine the type, amount, and location of marine debris in the bay. The mapping took place according to a prioritization scheme from Bergen municipality favoring certain areas within the 15m depth contour; as objects reaching 20cm over the seafloor within this focus area are required to be removed. The AROV system was deployed from the shoreline using a small mobile momentum crane, see Figure \ref{fig:survey}a. The mapping proceeded at 1.2m/s at around 2m altitude due to the high turbidity and low-light conditions, yielding an image swath of 2-3m. Figure \ref{fig:survey}b illustrates the described mapping procedure (UHI swath line shown), while Figure \ref{fig:survey}c shows the subsea lighting conditions more clearly during night time mapping. The seafloor was sequentially surveyed in blocks, where each mission block took from 20-40 minutes to survey. Data from the SCS and UHI was stored locally and uploaded to shore upon retrieval.  

\subsection{Marine debris detection results}
\begin{figure}[!h] 
\centering
\includegraphics[width=0.99\textwidth]{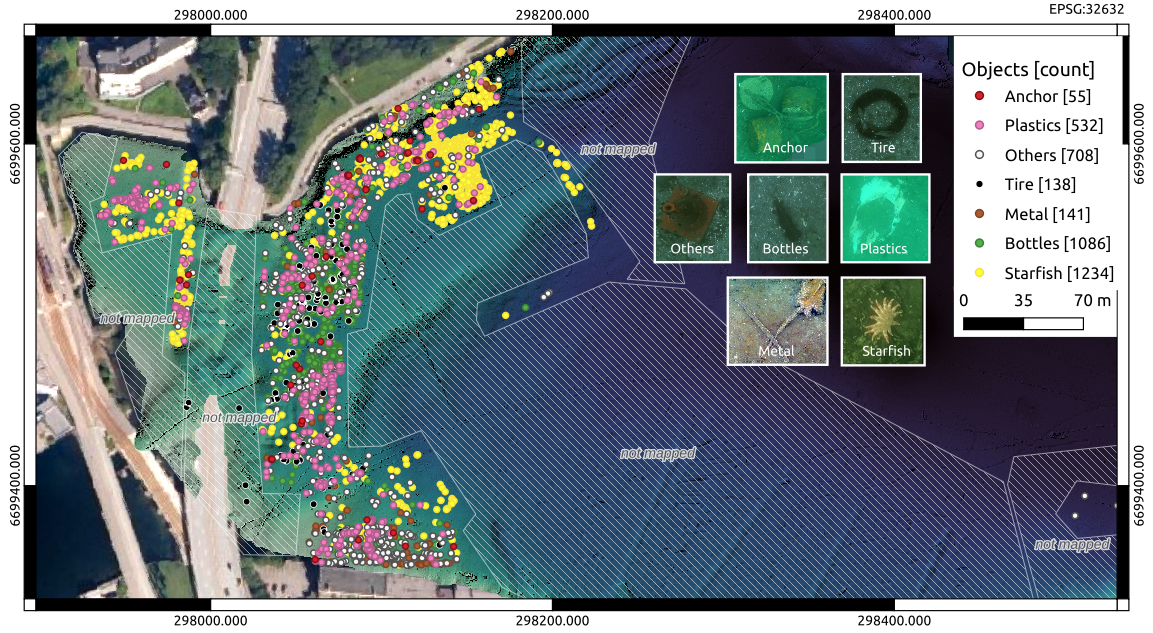}
\caption{Map over the western part of Store Lungegårdsvann showing the location of the detected objects and their classification.}
\label{fig:objmap}
\end{figure}

Using the ADS and CVI-tools a total of 3894 objects were found and classified into the following seven categories: bottles, plastics, anchors, tires, metal, and others. Starfish (Asteroidea) was also added as a detection group in order to give additional information about the local benthic epifauna. Manual inspection indicated that about 80-90\% of the objects were detected, depending on the turbidity conditions and sediment covering the objects. As expected, classes with many training examples had the best detection rate and vise-versa. The lower detection threshold used was set to 0.35 to avoid too many false positives and avoid laborious post-classification clean-up using the CVI-tool. A map of the detections is shown in Figure \ref{fig:objmap} along with color classification superimposed on multibeam bathymetry (shaded relief map). Certain areas were not mapped due to time limitation and prioritization, they are shown as white stripes with the label "not mapped". The distribution of the marine debris was different for each class. For example, plastic debris was found to be more evenly distributed compared to bottles and tires, which seemed to be a more concentrated in certain hot-spots. Aggregation of starfish can be seen in the northern and western parts of the mapped area, possibly tied to the local habitat. Using the detection location and a manually chosen average weight per class, the tonnage per hectare for each region was derived. The amount of debris varied from 400kg to over 10 tons per hectare for different regions in the bay, yielding highly relevant information for planning and structuring clean-up operations. 

\subsection{Photomosaics}
\begin{figure}[!h] 
\centering
\includegraphics[width=0.99\textwidth]{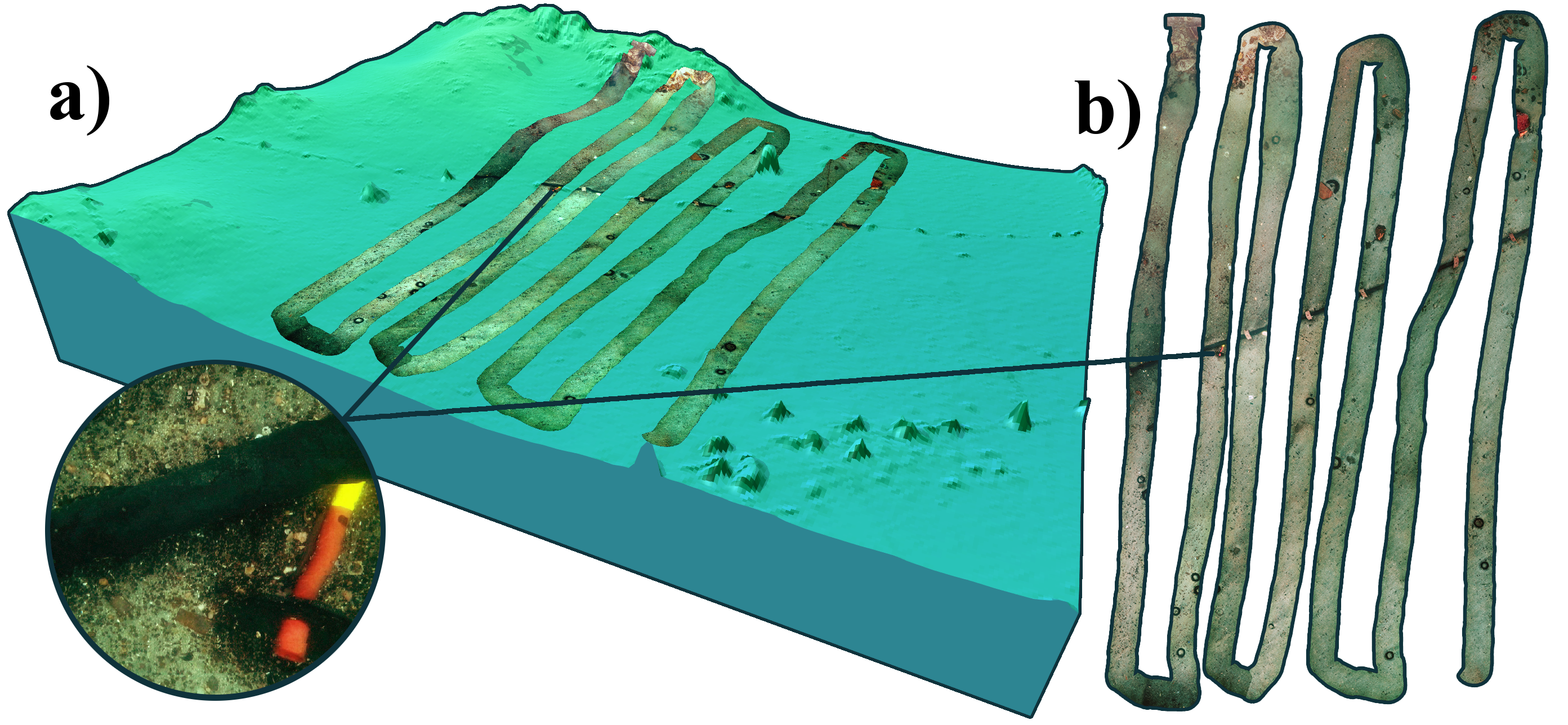}
\caption{a) Example 3D photomosaic from the survey draped on top of multibeam bathymetry. The resolution of the image mosaic is 0.5cm. b) 2D orthomosaic color corrected.}
\label{fig:ms10}
\end{figure}

Using the SCS a photomosaic representation of the survey can be produced. The mosaic-software, developed by Skarv Technologies$^1$ AS, uses image features, stereo information, and IMU data to stack, rotate, and scale (based on DVL height) the images sequentially to make a composite combining all the images ($\sim$5000) from the mission; attaining a resolution below 0.5cm. Figure \ref{fig:ms10}a shows the mosaic draped on the multibeam data, while Figure \ref{fig:ms10}b shows the mosaic for itself. Color correction was used to adjust the mosaic for attenuation and distance to the seafloor. A pipeline is visible running across the middle of the image, providing a good indirect measure of the estimated position. 

\subsection{Hyperspectral data}
\begin{figure}
     \centering
     \begin{subfigure}[!h]{0.99\textwidth}
         \centering
         \includegraphics[width=0.168\textwidth, angle=90]{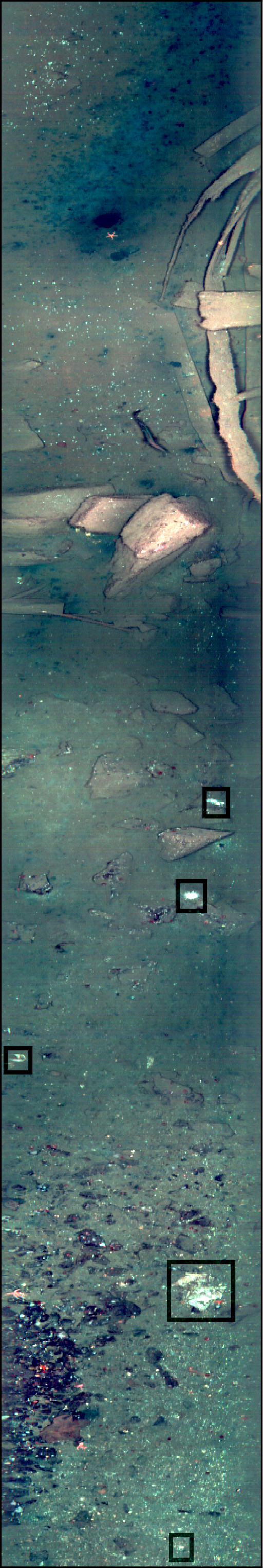}
         \caption{Pseudo-RGB representation and detection results.}
         \label{fig:uhia}
     \end{subfigure}
     \hfill
     \begin{subfigure}[!h]{0.99\textwidth}
         \centering
         \includegraphics[width=0.168\textwidth, angle=90]{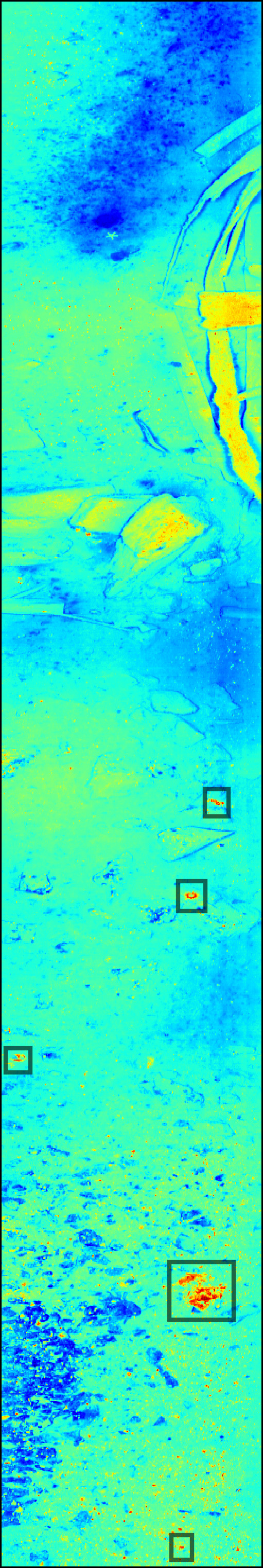}
         \caption{SAM-analysis using plastic reference spectrum and detection results.}
         \label{fig:uhib}
     \end{subfigure}
        \caption{Pseudo-RGB and SAM-analysis of a UHI-recording from the survey with detections of plastic-like materials. A partly covered plastic bag is visible in the larges bounding box. Parts of a shipwreck (protected cultural heritage) is also visible in the first part of the recording.}
        \label{fig:UHI}
\end{figure}

The UHI system  utilizes color information (spectrum) to compare and identify materials on the seafloor. For the purpose of this work the focus was on typical marine debris like plastics, metal, bottles, etc. Several types of debris materials were present on the seafloor, which prompted the use of unsupervised and semi-supervised methods as no prior library of spectral signatures (of typical marine debris) was available. A collection of spectra from the seabed and typical objects was therefore manually obtained using the UHI-recordings. Using these references the complete data set could be automatically analysed to look for similar or anomalous objects using different distance metrics (euclidean, cosine, etc.). Figure \ref{fig:UHI} presents results from spectral angle mapping (SAM) using a reference for white plastic. SAM is a standardized detection method for multi and hyperspectral analysis that uses cosine distance to measure the similarity of the data to a known reference. Selection of a similarity threshold can then be used to isolate objects that are similar (short cosine distance) or anomalous (large cosine distance). The first image in Figure \ref{fig:uhia} shows a pseudo-rgb representation (composite of 630nm, 532nm, and 465nm bands) of the hyperspectral recording together with bounding boxes from the automatic SAM-detection. The plastic bag is noticeable as the object with the biggest bounding box. The second image in Figure \ref{fig:uhib} shows the scene as a color graded SAM distance, where red is similar and blue is dis-similar materials. A simplified correction procedure was used to account for the distance to the seafloor (calibration plate) and local attenuation (measured with a separate instrument). The UHI-data was processed in combination with the stereo-images and fused in the CVI-tool, by adding to the existing feature vector. A more comprehensive fusion of the stereo and hyperspectral data is currently under development to exploit both high spatial- and spectral resolution in the context of object detection.   

\section{Discussion}
This work describes a methodology for autonomous mapping and detection of marine debris using a lightweight autonomous ROV equipped with stereo- and hyperspectral imaging sensors. The AROV system performed very well for this purpose, proving itself as a stable and controllable camera carrier. Despite being conducted in highly turbid waters the quality of the optical data was acceptable for doing both object detection and material classification, much due to adaptive survey parameters (speed and altitude of AUV) and specialized subsea lights. Exhaustive correction of inherent optical properties of the water and illumination effects on the UHI-data was not necessary for the analysis conducted here, as object detection was conducted in combination with the ADS. With that in mind, the UHI could have provided even more accurate results for targeted identification of certain material classes if this correction had been conducted. The ADS is a proof-of-concept and is not optimized for accuracy and performance. A relevant consideration in this regard is to evaluate different CNN-architectures like EfficientDet \cite{Wightman}, which have been successfully used by other relevant work for detection of marine debris \cite{MAJCHROWSKA2022274, zocco2022towards}. The incorporation of more specific marine debris classes would be possible. Addressing this problem would require a substantially larger data set that accounts for the wide range of debris, degradation, morphology, and environmental conditions that can arise. More analysis could be done to better exploit the synergies between stereo and hyperspectral data to mitigate this problem, by employing better co-registration methods like described in \cite{lovaas2021methodology}.

\section{Conclusion}
Based on the results from this survey, autonomous systems and machine learning, combined with optical instruments like stereo- and hyperspectral cameras, have shown to provide an effective and useful tool for mapping marine debris. Using pattern and spectral features allow detection schemes that are more robust and accurate when surveying in turbid and sedimented areas. Challenges related to these environmental factors, as well as pattern variation (e.g. degraded plastic) and material diversity is still a problem that will require further development.    

\section*{Acknowledgements}
Skarv Technology AS acknowledges support from Handelens Miljøfond\footnote{\url{https://handelensmiljofond.no/}} project number 12458, and the \texttt{ATOM} project, funded by the Research Council of Norway (project number 321592). The authors are also grateful for the support and help from the staff in Bergen City Agency for Urban Environment, The AUR-Lab at the Norwegian University of Science and Technology (NTNU), and the NOSCA Clean Oceans cluster\footnote{\url{https://www.nosca.no/}}.
\printbibliography

\end{document}